\documentclass[twoside,11pt]{article}

\usepackage{jmlr2e}
\usepackage{amsmath}
\usepackage{amssymb}
\usepackage{caption}
\usepackage{bm} 
\usepackage{url}
\usepackage{natbib}
\usepackage{multirow}
\usepackage{subfigure}
\usepackage{makecell}
\usepackage{booktabs}
\usepackage{array}
\usepackage{url}
\usepackage{algorithm}
\usepackage{algorithmic}
\usepackage{smile}
\usepackage{psfrag}
\usepackage[format=hang]{caption}
\usepackage[colorlinks=false,allbordercolors={1 1 1}]{hyperref}

\jmlrheading{16}{2015}{553-557}{11/12; Revised 3/14}{3/15}{Xingguo Li, Tuo Zhao, Xiaoming Yuan and Han Liu}

\ShortHeadings{The \texttt{flare} Package in \texttt{R}}{Li, Zhao, Yuan and Liu}
\firstpageno{1}

\begin{document}

\title{The \texttt{flare} Package for High Dimensional Linear Regression and Precision Matrix Estimation in \texttt{R}\thanks{The package vignette is an extended version of this paper, which contains more technical details.}}

\author{\name Xingguo Li\thanks{Xingguo Li and Tuo Zhao contributed equally to this work.} \email lixx1661@umn.edu\\
	\addr Department of Electrical and Computer Engineering\\
       		 University of Minnesota Twin Cities\\
       		 Minneapolis, MN, 55455, USA\\
	\name Tuo Zhao$^\dagger$ \email tzhao5@jhu.edu \\
       	\addr Department of Computer Science\\
       		 Johns Hopkins University\\
       		 Baltimore, MD, 21210, USA\\
       	\name Xiaoming Yuan \email  xmyuan@hkbu.edu.hk\\
       	\addr Department of Mathematics\\
		 Hong Kong Baptist University\\
		 Hong Kong, China\\
       	\name Han Liu \email hanliu@princeton.edu\\
       	\addr Department of Operations Research and Financial Engineering\\
		 Princeton University\\
		 Princeton, NJ 08544, USA}

\editor{Mikio Braun}

\maketitle

\begin{abstract}
This paper describes an \texttt{R} package named \texttt{flare}, which implements a family of new high dimensional regression methods (LAD Lasso, SQRT Lasso, $\ell_q$ Lasso, and Dantzig selector) and their extensions to sparse precision matrix estimation (TIGER and CLIME). These methods exploit different nonsmooth loss functions to gain modeling flexibility, estimation robustness, and tuning insensitiveness. The developed solver is based on the alternating direction method of multipliers (ADMM). The package \texttt{flare} is coded in double precision \texttt{C}, and called from \texttt{R} by a user-friendly interface. The memory usage is optimized by using the sparse matrix output. The experiments show that \texttt{flare} is efficient and can scale up to large problems.
\end{abstract}


\section{Introduction}

As a popular sparse linear regression method for high dimensional data analysis, Lasso has been extensively studied by machine learning scientists \citep{tibshirani96Lasso}. It adopts the $\ell_1$-regularized least square formulation to select and estimate nonzero parameters simultaneously. Software packages such as \texttt{glmnet} and \texttt{huge} have been developed to efficiently solve large problems \citep{friedman10glmnet,zhao2012huge,zhao2014general}. Lasso further yields a wide range of research interests, and motivates many variants by exploiting nonsmooth loss functions to gain modeling flexibility, estimation robustness, and tuning insensitiveness (See more details in the package vignette, \cite{zhao2014calibrated,liu2014multivariate}). These nonsmooth loss functions pose a great challenge to computation. To the best of our knowledge, no efficient solver has been developed so far for these Lasso variants.

In this report, we describe a newly developed \texttt{R} package named \texttt{flare} (\underline{F}amily of \underline{La}sso \underline{Re}gression). The \texttt{flare} package implements a family of linear regression methods including: (1) LAD Lasso, which is robust to heavy tail random noise and outliers \citep{wang2013l1}; (2) SQRT Lasso, which is tuning insensitive (the optimal regularization parameter selection does not depend on any unknown parameter, \cite{belloni11sLasso}); (3) $\ell_q$ Lasso, which shares the advantage of LAD Lasso and SQRT Lasso; (4) Dantzig selector, which can tolerate missing values in the design matrix and response vector \citep{candes07dantzig}. By adopting the column by column regression scheme, we further extend these regression methods to sparse precision matrix estimation, including: (5) TIGER, which is tuning insensitive \citep{liu12tiger}; (6) CLIME, which can tolerate missing values in the data matrix \citep{cai11clime}. The developed solver is based on the alternating direction method of multipliers (ADMM), which is further accelerated by a multistage screening approach \citep{boyd2011distributed,liu2014sparse}. The global convergence result of ADMM has been established in \cite{he12nonergodic,he12ergodic}. The numerical simulations show that the \texttt{flare} package is efficient and can scale up to large problems.

\section{Algorithm}

We are interested in solving convex programs in the following generic form
\begin{align}\label{ADMM}
\hat{\bbeta} =\argmin_{\bbeta,~\balpha} L_\lambda(\balpha) +\|\bbeta\|_1~~~\textrm{subject to}~\br -\Ab\bbeta =\balpha.
\end{align}
where $\lambda>0$ is the regularization parameter. The possible choices of $L_\lambda(\balpha)$, $\Ab$, and $\br$ for different regression methods are listed in Table \ref{choice}. Note that LAD Lasso and SQRT Lasso are special examples of $\ell_q$ Lasso for $q=1$ and $q=2$ respectively.

\noindent All methods in Table \ref{choice} can be efficiently solved by the iterative scheme as follows
\begin{align}
\balpha^{t+1} &=\argmin_{\balpha}\frac{1}{2}\left\|\bu^t+\br-\Ab\bbeta^{t}-\balpha\right\|_2^2 +\frac{1}{\rho}L_{\lambda}(\balpha),\label{update1}\\
\bbeta^{t+1} &=\argmin_{\bbeta}\frac{1}{2}\left\|\bu^t -\balpha^{t+1} +\br -\Ab\bbeta\right\|_2^2 +\frac{1}{\rho}\|\bbeta\|_1,\label{update2}\\
\bu^{t+1} &=\bu^{t} + (\br -\balpha^{t+1} -\Ab\bbeta^{t+1}),\label{update3}
\end{align}
where $\bu$ is the rescaled Lagrange multiplier \citep{boyd2011distributed}, and $\rho>0$ is the penalty parameter. For LAD Lasso, SQRT Lasso, or Dantzig selector, \eqref{update1} has a closed form solution via the winsorization, soft thresholding, and group soft thresholding operators respectively. For $L_q$ Lasso with $1<q<2$, \eqref{update1} can be solved by the bisection-based root finding algorithm. \eqref{update2} is a Lasso problem, which can be (approximately) solved by linearization or coordinate descent. Besides the pathwise optimization scheme and the active set trick, we also adopt the multistage screening approach to speedup the computation. In particular, we first select $k$ nested subsets of coordinates $\cA_1\subseteq\cA_2\subseteq...\subseteq\cA_k=\RR^d$ by the marginal correlation between the covariates and responses. Then the algorithm iterates over these nested subsets of coordinates to obtain the solution. The multistage screening approach can greatly boost the empirical performance, especially for Dantzig selector.

\begin{table}[htb!]
\centering
\begin{tabular}[t]{ c | c | c | c | c }
\Xhline{1pt}
Method &Loss function &$\Ab$ &$\br$ &Existing solver\\
\Xhline{1 pt}
\multirow{2}{*}{$L_q$ Lasso} &\multirow{2}{*}{$\displaystyle L_\lambda(\balpha)=\frac{1}{\sqrt[q]{n}\lambda}\| \balpha \|_q$}&\multirow{2}{*}{$\Xb$} &\multirow{2}{*}{$\by$} &\multirow{2}{*}{L.P. or S.O.C.P.}\\
&&&&\\
\hline
\multirow{3}{*}{Dantzig selector} &\multirow{3}{*}{$\displaystyle L_\lambda(\balpha)= \left \{ 
\begin{array}{ll}
\infty & \textrm{if $\| \balpha \|_{\infty} > \lambda$}\\
0 &\textrm{otherwise}
\end{array}\right.$
}&\multirow{3}{*}{$\frac{1}{n}\Xb^T\Xb$} &\multirow{3}{*}{$\frac{1}{n}\Xb^T\by$} &\multirow{3}{*}{L.P.}\\
&&&&\\
&&&&\\               
\Xhline{1pt}
\end{tabular}
\caption{All regression methods provided in the \texttt{flare} package. $\Xb\in\RR^{n\times d}$ denotes the design matrix, and $\by\in\RR^n$ denotes the response vector.  ``L.P." denotes the general linear programming solver, and ``S.O.C.P" denotes the second-order cone programming solver.}
\label{choice}
\end{table}

\section{Examples}

We illustrate the user interface by analyzing the eye disease data set in \texttt{flare}.
\begin{verbatim}
> # Load the data set
> library(flare); data(eyedata)
> # SQRT Lasso
> out1 = slim(x,y,method="lq",nlambda=40,lambda.min.value=sqrt(log(200)/120))
> # Dantzig Selector
> out2 = slim(x,y,method="dantzig",nlambda=40,lambda.min.ratio=0.35)
\end{verbatim}
The program automatically generates a sequence of 40 regularization parameters and estimates the corresponding solution paths of SQRT Lasso and the Dantzig selector. For the Dantzig selector, the optimal regularization parameter is usually selected based on some model selection procedures, such as cross validation. Note that \cite{belloni11sLasso} has shown that the theoretically consistent regularization parameter of  SQRT Lasso is $C\sqrt{\log d}/n$, where $C$ is some constant. Thus we manually choose its minimum regularization parameter to be $\sqrt{\log(d)/n}=\sqrt{\log(200)/120}$. The minimum regularization parameter yields 19 nonzero coefficients out of 200.

\section{Numerical Simulation}

All experiments below are carried out on a PC with Intel Core i5 3.3GHz processor, and the convergence threshold of \texttt{flare} is chosen to be $10^{-5}$. Timings (in seconds) are averaged over 100 replications using 20 regularization parameters, and the range of regularization parameters is chosen so that each method produces approximately the same number of nonzero estimates.

We first evaluate the timing performance of \texttt{flare} for sparse linear regression. We set $n=100$ and vary $d$ from 375 to 3000 as is shown in Table \ref{comparison-table}. We independently generate each row of the design matrix from a $d$-dimensional normal distribution $N(0,\bSigma)$, where $\bSigma_{jk} = 0.5^{|j-k|}$. Then we generate the response vector using $y_i = 3\Xb_{i1}+2\Xb_{i2}+1.5\Xb_{i4} +\epsilon_i$, where $\epsilon_i$ is independently generated from $N(0,1)$. From Table \ref{comparison-table}, we see that all methods achieve good timing performance. Dantzig selector and $\ell_q$ Lasso are slower than the others due to more difficult computational formulations.

We then evaluate the timing performance of \texttt{flare} for sparse precision matrix estimation. We set $n=100$ and vary $d$ from 100 to 400 as is shown in Table \ref{comparison-table}. We independently generate the data from  a $d$-dimensional normal distribution $N(0,\bSigma)$, where $\bSigma_{jk} = 0.5^{|j-k|}$. The corresponding precision matrix $\bOmega =\bSigma^{-1}$ has $\bOmega_{jj} = 1.3333$, $\bOmega_{jk} = -0.6667$ for all $j,k=1,...,d$ and $|j-k|=1$, and all other entries are 0. From Table \ref{comparison-table}, we see that both TIGER and CLIME achieve good timing performance, and CLIME is slower than TIGER due to a more difficult computational formulation.

\begin{table}[htp!]
\centering
\begin{tabular}[t]{l | l | c | c | c | c | c}
\Xhline{1pt}
\multicolumn{7}{c}{Sparse Linear Regression}\\
\hline
\multicolumn{3}{c|}{Method} & $d=375$ &$d=750$ &$d=1500$ &$d=3000$\\
\Xhline{1 pt}
\multicolumn{3}{l|}{LAD Lasso}&1.1713(0.2915)&1.1046(0.3640) &1.8103(0.2919) &3.1378(0.7753)\\
\multicolumn{3}{l|}{SQRT Lasso} &0.4888(0.0264)&0.7330(0.1234)&0.9485(0.2167) &1.2761(0.1510)\\
\multicolumn{3}{l|}{$\ell_{1.5}$ Lasso} &12.995(0.5535)&14.071(0.5966)&14.382(0.7390) &16.936(0.5696)\\
\multicolumn{3}{l|}{Dantzig selector} &0.3245(0.1871)&1.5360(1.8566)&4.4669(5.9929) &17.034(23.202)\\
\Xhline{1pt}
\multicolumn{7}{c}{Sparse Precision Matrix Estimation}\\
\hline
\multicolumn{3}{c|}{Method} & $d=100$ &$d=200$ &$d=300$ &d=400\\
\Xhline{1 pt}
\multicolumn{3}{l|}{TIGER} &1.0637(0.0361)&4.6251(0.0807)&7.1860(0.0795)&11.085(0.1715)\\
\multicolumn{3}{l|}{CLIME} &2.5761(0.3807) &20.137(3.2258) &42.882(18.188) &112.50(11.561)\\
\Xhline{1pt}
\end{tabular}
\caption{Average timing  performance (in seconds) with standard errors in the parentheses on sparse linear regression and sparse precision matrix estimation.}
\label{comparison-table}
\end{table}

\section{Discussion and Conclusions}

Though the \texttt{glmnet} package cannot handle nonsmooth loss functions, it is much faster than \texttt{flare} for solving Lasso,\footnote{See more detail in the package vignette.} and the \texttt{glmnet} package can also be applied to solve $\ell_1$ regularized generalized linear model estimation problems, which \texttt{flare} cannot. Overall speaking, the \texttt{flare} package serves as an efficient complement to the \texttt{glmnet} package for high dimensional data analysis. We will continue to maintain and support this package.

\acks{Tuo Zhao and Han Liu are supported by NSF Grants III-1116730 and NSF III-1332109, NIH R01MH102339, NIH R01GM083084, and NIH R01HG06841, and FDA HHSF2232 01000072C. Xiaoming Yuan is supported by the General Research Fund form Hong Kong Research Grants Council: 203311 and 203712.}

\bibliography{Library}
\end{document}